# Improved Image Deblurring based on Salient-region Segmentation

Chongyang Zhang[1], Weiyao Lin[1+], Wei Li[2], Bing Zhou[3], Jun Xie[1], Jijia Li[1]

[1] Department of Electronic Engineering, Shanghai Jiao Tong University, Shanghai 200240, China

[2] The State Key Laboratory of Virtual Reality Technology and Systems, School of Computer Science and Engineering, Beihang University, Beijing 100191, China

[3] School of Information Engineering, Zhengzhou University, Zhengzhou 450001, China

([+] Corresponding author)

**Abstract**

Image deblurring techniques play important roles in many image processing applications. As the blur varies spatially across the image plane, it calls for robust and effective methods to deal with the spatially-variant blur problem. In this paper, a Saliency-based Deblurring (SD) approach is proposed based on the saliency detection for salient-region segmentation and a corresponding compensate method for image deblurring. We also propose a PDE-based deblurring method which introduces an anisotropic Partial Differential Equation (PDE) model for latent image prediction and employs an adaptive optimization model in the kernel estimation and deconvolution steps. Experimental results demonstrate the effectiveness of the proposed algorithm.

**Index Terms** — Image Deblurring, Salient-region Segmentation, Motion Deblurring.

# I. INTRODUCTION AND RELATED WORKS

Normally, motion blurs are produced when the object image changes during the recording of a single frame, either due to rapid movement or long exposure. The undesired blurry photograph with inevitable information loss may not only have a bad visualization but also result in a degrading effect in further process such as feature extraction or object recognition. Therefore, it is always desirable to develop efficient algorithms to remove these motion blurs [1]-[4]. Specifically, it is of increasing importance for many consumer devices (such as digital camera, tablet PC, or surveillance devices) to have deblurring capability to handle the possible motion blurs.

The motion blur can be formulated as the convolution of a latent image with a blur kernel. When the kernel is unknown, removing the blur from an image is thus a blind-deconvolution operation, which involves lots of challenges in modeling and optimization. Early algorithms only focus on parameterized kernel forms or small size kernels [2]-[4], which made the result less desirable. Some researchers use more than one blurry image to estimate the blur kernel [5]-[6]. Although these methods can produce excellent deblurring results, sometimes finding multiple images is difficult and it cannot be easily used in every scenario.

Hence, single-image blind deconvolution has become the focus of recent deblurring researches [7]-[14]. This ill-posed problem can be mainly divided into two phases: kernel estimation and deconvolution. In kernel estimation, sparse priors are widely adopted. Fergus et al. [7] uses a variational Bayesian method with Mixture of Gaussian to estimate natural image statistics. However, this algorithm needs intensive computation. Furthermore, Shan et al. [8] proposes a novel scheme with consideration of the random noise model to reduce ringing artifacts. Besides, this method uses an efficient optimization constraint that alternates between blur kernel estimation and latent image restoration. However, this method is also time-consuming and sometimes the noises cannot be effectively suppressed. Some other researchers use edge prediction in the kernel estimation step [9]-[11]. To speed up the deblurring process, Cho et al. [9] presents a new method by introducing a derivative-based edge prediction step. This scheme uses a combination of bilateral filter and shock filter to predict the latent image's edge in iterative multiple scales and performs fast kernel estimation and image deconvolution under optimal constraints alternatively. Although this method has less complexity, its results are vulnerable to noises. Recently, Xu et al. [11] employs iterative support detection (ISD) for kernel estimation as well as a robust TV-$l_1$ algorithm with large Point Spread Function (PSF) for deconvolution. Hu et al. [12] perform deblurring by extracting "good regions" within the Conditional Random Field framework. These approaches improve the deblurring effect. However, they may still lead to some unnatural effects in the result.

In some algorithms, only image restoration process is discussed, which simplifies the problem into a non-blind deconvolution scheme. Early works such as Weiner filtering and Richardson-Lucy belong to this category. Yuan et al. [13] proposes an effective edge-preserving non-blind deconvolution approach and significantly reduces ringing artifacts. Therefore, if kernel prediction and robust deconvolution can be effectively combined, the deblurring effects will be greatly enhanced. Furthermore, some hardware approaches have also been implemented in the motion deblurring problems [15]-[16]. For example, Joshi et al. [16] uses a hardware attachment coupled with a natural image prior to deblur images from consumer cameras. However, these methods need specific system requirements and thus are not applicable on many of the existing consumer devices.

On the other hand, for some blurry images, when the captured object is away from the focus plane, it may cause spatially-variant blur problems. For example, the foreground object is in focus whereas the background information is blurred. In these cases, the blur is distributed unevenly and if general deblurring methods are applied, not only the local blur cannot be removed effectively but they may produce undesired artifacts in the originally-sharp areas either, as will be discussed later. Levin et al. [21] manages to retrieve coarse depth information by modifying the conventional camera system. This approach can robustly realize refocusing and depth-based image editing. However, their method has specific assumptions and hardware requirements which limit their applications. Cho et al. [17] proposes an algorithm for removing spatially varying motion blurs from multiple images. The disadvantage of this method is that at least two input images are needed to analyze the spatially variant blur. Recently, Chakrabarti et al. [18] proposes a novel way for analyzing spatially-varying blur from a single image. Although this algorithm is robust, it only focuses on foreground blurry object detection and extraction. Furthermore, Chan et al. [19] also addresses the local blur problem by using background prediction. However, this method still needs further improvements under more complicated scenarios. Thus, it is desirable to develop a general algorithm which can handle spatially variant blur problems with robust deblurring approaches.

In this paper, a new algorithm is proposed for motion deblurring. The algorithm addresses the spatially variant blur problem by proposing a Saliency-based Deblurring (SD) method and a compensate approach such that the blur is locally removed whereas sharp components are preserved. Furthermore, we also propose a PDE-based deblurring method which adopts anisotropic PDE model for edge prediction as the initial step to estimate PSF and employs an adaptive optimization constraint for kernel estimation and deconvolution based on image derivatives. The kernel estimation and latent image restoration are performed in multi-scale iteration until convergence. Experimental results show that our proposed algorithm has both satisfactory deblurring performances and low computation complexity. Thus, the proposed algorithm is suitable to be applied on various consumer devices for handling the motion blur problems.

The rest of the paper is organized as follows: Section II presents our proposed deblurring algorithm. The saliency-based deblurring approach and the PDE model based adaptive deconvolution are described with details. Section III shows the experimental results and Section IV concludes the paper.

## II. THE PROPOSED DEBLURRING METHOD

### A. The Framework

For many motion blur cases, the blur effect may not be uniform in the entire image because of the various depths of recorded objects, selective focuses, or camera rotation. Therefore, spatially-variant blur becomes an inevitable problem in the deblurring research domain. Given an image with spatially-variant blur, we cannot simply apply the current deblurring approaches to this case due to the following reasons: (1) Different parts in the image have different blur degrees. Some may be very sharp whereas others may have various blur directions. Therefore, it is impossible to estimate a uniform blur kernel for the entire image. (2) Even if we can successfully estimate the kernel for the blurry part in the image, deconvolution will also cause undesired artifacts around the component which is originally sharp. Therefore, if we can estimate the blur kernel partially and perform deconvolution throughout the entire image without introducing artifacts, the spatially-variant blur problem can be handled properly.

For the ease of description, we will discuss our idea based on the case where the foreground image is sharp whereas the background parts are uniformly blurred due to motion blur or defocus. It should be noted that our idea is general and it can be extended to other blurry cases with slight modifications.

According to this problem, our first attempt is to extract the foreground region from the blurred background. Here we propose to use the saliency method as visual saliency is the perceptual quality that makes an object stand out relative to its neighboring items. If the foreground region is sharp, it will be easily recognized as a salient part from the background. In our method, we adopt the Frequency-tuned Salient Region Detection algorithm proposed by Achanta et al. [20] for saliency detection. Due to the limited space, we will not discuss the saliency detection process in detail and only show its results in Fig. 1. Fig. 1 is an example illustrating the saliency detection result by Achanta's method [20]. We can see from Fig. 1 that most of the sharp regions can be effectively segmented by the detected saliency features. It should be noted that saliency detection is just one of effective methods to separate blurry part and sharp part from the entire image. Apart from saliency detection, more complex approaches can be introduced to deal with the blurry part extraction [18]-[19]. Besides, the image segmentation methods such as graph cut [26] can also be used for saliency region segmentation. This point will be further discussed in Section III-C later.

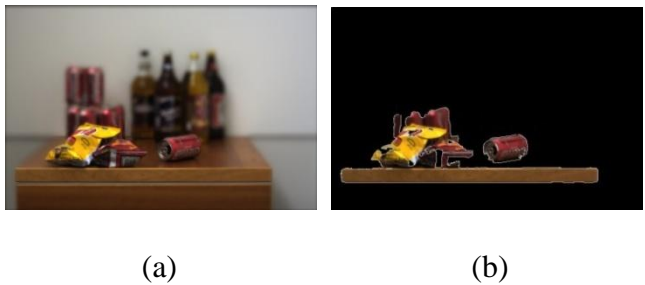

(a)          (b)

Fig. 1. Saliency detection result. (a) the original spatially-invariant blur image, (b) the segmented saliency part.

The framework of our proposed Saliency-based Deblurring algorithm is illustrated in Fig. 2. Our aim is to deblur the rest background part while keeping the foreground components sharp and natural. In the first step, we separate the foreground and background components by the saliency map and extract a largest rectangle in the background area which does not include the saliency part. Then we perform the kernel estimation to this rectangular area and get a blur kernel $K$. As the background blur is uniform, we can conclude that the blur kernel applies to all the background blurry area. For the deconvolution step, we adopt a compensate method to preprocess the original image such that a uniform blur is distributed throughout the entire image and the blur kernel can be applied globally for deconvolution. In our algorithm, we convolve $K$ with the segmented saliency part and fuse this blurred convolution result with the background together. Hence, the fusion result becomes a uniform blurry image with the same blur kernel globally. Then deconvolution is performed to the entire modified image by using the background blur kernel $K$ estimated before. If we use the blur kernel for deconvolution without preprocessing, undesired artifacts will be introduced around the boundaries between the blurry and sharp parts even though we keep the sharp part unchanged. The detailed description of the steps in Fig. 2 is described in the following.

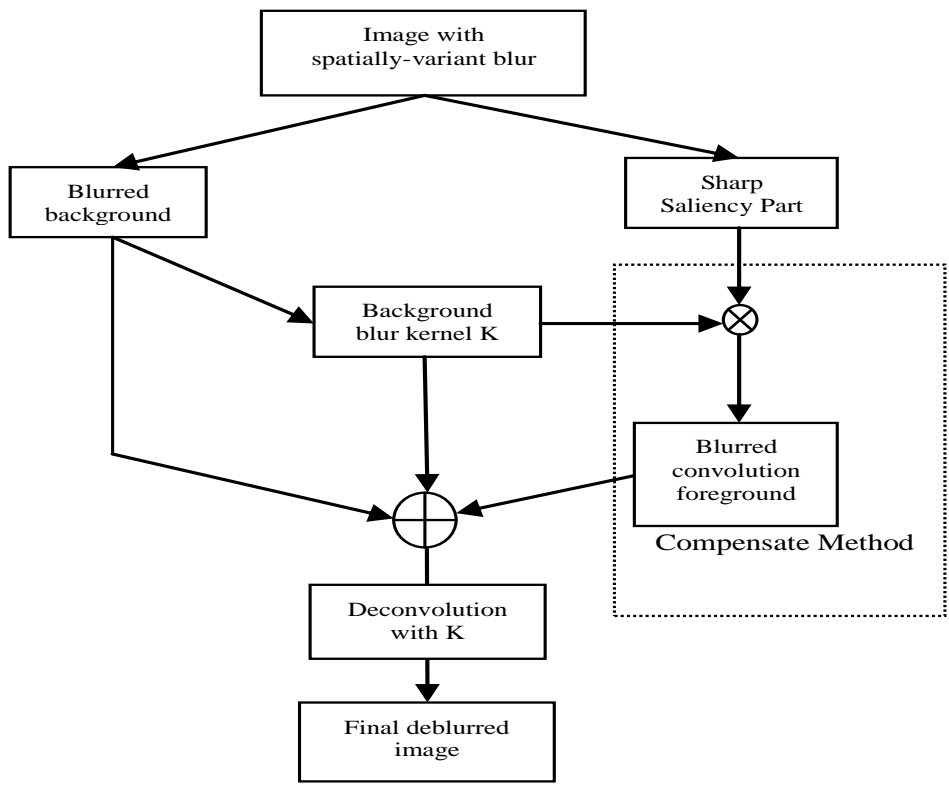

Fig. 2. The framework of the Saliency-based Deblurring algorithm.

In the separation step, we use the binary saliency map mask (*MAP*) where white indicates the saliency position and black corresponds to the background components. To remove boundary artifacts, the saliency part in the map is dilated. Therefore, the saliency and background components can be formulated as:

$$\begin{aligned} Saliency &= MAP \cdot Original\_image \\ Background &= (1 - MAP) \cdot Original\_image \end{aligned} \quad (1)$$

In the compensate method, the fused image with uniform blur is modeled as:

$$Fused\_image = MAP \cdot (K \otimes Original\_image) + Background \qquad (2)$$

In the final step, after we get the final deblurred image from the fused image in (2), we can fuse it again with the sharp saliency part in a similar way such that the result can be more accurate. That is,

$$Final = MAP \cdot Original\_image + (1 - MAP) \cdot Deblurred\_image \qquad (3)$$

Furthermore, the steps of "background blur kernel $K$ estimation" and "deconvolution with $K$" in Fig. 2 are implemented by our proposed PDE-based deblurring method. It is described in the following.

### B. PDE-based Deblurring Method

As mentioned, most existing motion deblurring methods still have limitations in artifacts suppressions. We observe that in many cases, these undesired artifacts are mainly caused by inaccurate Point Spread Function (PSF) estimation or edge prediction as well as inappropriate optimization modeling in the deconvolution step. Therefore, new algorithms which can improve these models are needed.

In our PDE-based deblurring method, the deblurring process is modeled iteratively where the steps of latent image prediction, kernel estimation, and deconvolution are alternating in a coarse-to-fine scheme. After a suitable kernel $K$ is estimated, an additional deconvolution step is performed to achieve the final deblurred image. The framework of the proposed deblurring algorithm can be described as in Fig. 3 and the above three main steps are described in details in the following. It should be noted that our PDE-based deblurring method can be directly used for image deblurring. When it is included in our Saliency-based Deblurring framework, the iterative parts (i.e., the grey blocks in Fig. 3) are used for background blur kernel *K* estimation and the additional deconvolution step (i.e., the dashed block in Fig. 3) is used for the step "deconvolution with *K*".

*1) Latent Image Prediction*

In the first place, edge prediction is performed for predicting a latent image, which is used as an initial input of kernel estimation. In [9], shock filter is introduced to pre-sharp the blurry image. But it may lead to some undesired artifacts caused by the "double edges" around the true edge as well as inevitable information loss in the predicted latent image. In our method, we propose to use a bilateral filter to pre-smooth the input image first and then solve an anisotropic PDE [21]-[23] to enhance the image's true edges. The description of anisotropic PDE is described in the following:

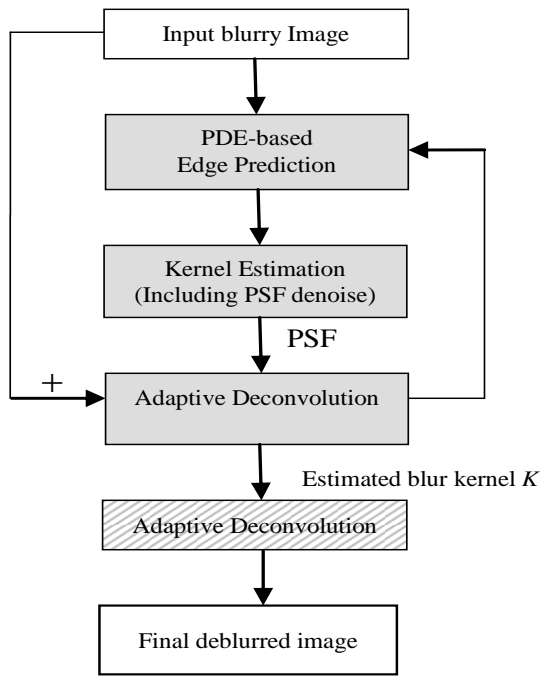

Fig. 3. The proposed iterative deblurring algorithm.

The PDE of a 2D scalar image *I* can be formulated as the juxtaposition of two 1D flows along the gradient direction $\xi$ and its orthogonal direction $\eta$, that is,

$$\frac{\partial I}{\partial t} = c_\xi \frac{\partial^2 I}{\partial \xi^2} + c_\eta \frac{\partial^2 I}{\partial \eta^2} \quad \eta = \frac{\nabla I}{\|\nabla I\|}, \xi = \eta^\perp \qquad (4)$$

where $\|\nabla I\|$ denotes the image gradient magnitude. $c_\xi$ and $c_\eta$ are the corresponding weights suggested in [21]-[22]:

$$c_\eta = \frac{1}{1+\|\nabla I\|^2}, \quad c_\xi = \frac{1}{\sqrt{1+\|\nabla I\|^2}}, \qquad (5)$$

The PDE in (4) can be solved as

$$\frac{\partial I}{\partial t} = trace(TH) = \Delta I \qquad (6)$$

where $H$ is the Hessian matrix of $I$ and $T$ is a 2×2 tensor defined as:

$$H = \begin{bmatrix} I_{xx} & I_{xy} \\ I_{xy} & I_{yy} \end{bmatrix} \text{ and } T = c_\xi \xi \xi^T + c_\eta \eta \eta^T \qquad (7)$$

In order to enhance edges, we update the predicted latent image as:

$$I_{pred} = I - \lambda \Delta I \qquad (8)$$

where $\lambda$ is a controlling parameter which varies in each iteration at one scale.

In the anisotropic PDE process, $-\lambda \Delta I$ represents the true edge map of the blurry image. Therefore, the purpose of predicted latent image update is to enhance the true edges of the blurry object. Results by our method have shown that the true edges are enhanced whereas the image nature is also preserved. Thus the edge prediction can be more accurate for the consequential kernel estimation.

*2) Kernel Estimation*

In the kernel estimation step, Point Spread Function (PSF) is estimated from the blurry image $B$ and the gradient maps $P_*$ of the previously predicted latent image $L$ ($L$ is the desired deblurred image). And the resulting PSF will become the estimated blur kernel $K$. Basically, the blur kernel $K$ can be estimated such that the latent image $L$ is similar to the blurry image $B$ when convolved with $K$:

$$K = \arg\min_{K'} \{\|B - K' \otimes L\| + \rho_K(K')\} \tag{9}$$

where $\otimes$ is the convolution operator, and $\rho_K(K)$ is the regularization term.

By using the gradient maps $P_*$, (9) can be solved by minimizing the energy function $f(K)$ in the following [8], [9], [11]:

$$K = \arg\min(f(K)) \tag{10}$$

$$f(K) = \sum_{(P_*, \partial_* B)} w_* \|K \otimes P_* - \partial_* B\|^2 + \theta \|K\|^2$$

where $(P_*, \partial_* B) \in \{(P_x, \partial_x B), (P_y, \partial_y B), (\partial_x P_x, \partial_{xx} B),$ $(\partial_y P_y, \partial_{yy} B), ((\partial_x P_y + \partial_y P_x)/2, \partial_{xy} B)\}$ and $P_* = (P_x, P_y)$ is the threshold-based gradient map of $L$ [9]. The purpose of using threshold-based gradient map is to eliminate the effects of double edges. $\partial_i$ is the derivation operation on $i$ direction. $\otimes$ is the convolution operation. $\theta$ is the weight for the Tikhonov Regularization and $w_*$ includes the weights for each partial derivatives. In our experiment, $w_*$ is set to be {25, 25, 12.5, 12.5, 12.5} [9] and $\theta$ equals 5. Note that the first term in $f(K)$ can be viewed as the derivative version of the first term in (9) while the second term in $f(K)$ corresponds to the regularization term in (9). The solution for (10) is [8, 9]:

$$F(K) = \frac{\sum_{(P_*, \partial_* B)} w_* \cdot \overline{F(P_*)} \cdot F(\partial_* B)}{\sum_{(P_*, \partial_* B)} w_* \cdot \overline{F(P_*)} \cdot F(P_*) + \gamma \cdot I_B} \tag{11}$$

where $F(*)$ denotes the Fourier Transformation, $\overline{F(*)}$ is the conjugate of the Fourier Transformation, and $I_B$ is an all-one matrix which has the same size with the blurry image $B$, and $\gamma$ is a weighting parameter which is set to be 5 in our algorithm.

However, when PSF is estimated by Fourier transformation, noises may be produced especially after the predicted image $L$ is up-sampled in the multi-scale scenario. Therefore, we further propose to perform a denoising process for the blur kernel after PSF estimation, that is, first let the pixels to be zero if their values are less than 1/20 of the maximum pixel value. And then remove the unconnected components whose area is less than $1/D$ of the kernel size. Here $D$ is determined by the parameters during each iteration. For simplicity, it can be chosen from the interval between 128 and 256. This process can be viewed as deleting those non-good points during kernel estimation. After normalization, the refined PSF is our estimated blur kernel in one iteration.

*3) Adaptive Deconvolution*

With the estimated kernel in the previous step, our basic deconvolution idea is to restore the latent image $L$ by minimizing the energy function $f(L)$ from the estimated kernel $K$ and the input blurry image $B$. Basically, the latent image $L$ is estimated such that $L$ is similar to $B$ when convolved with $K$:

$$L = \arg\min_{L'}\{\|B - K \otimes L'\| + \rho_L(L')\} \tag{12}$$

where $\otimes$ is the convolution operator, and $\rho_L(L)$ is the regularization term. Similar to (10), the problem of (12) can be solved by the gradient information as:

$$L = \arg\min(f(L)) \tag{13}$$

$$f(L) = \sum_{\partial_*} \omega_* \|K \otimes \partial_* L - \partial_* B\|^2 + \alpha \|\nabla L - v\|^2 + \beta \|v\|$$

where $\partial_* \in \{\partial_0, \partial_x, \partial_y, \partial_{xx}, \partial_{yy}, \partial_{xy}\}$, $\omega_*$ is the corresponding weights which is set similar to $w_*$: {50, 25, 25, 12.5, 12.5, 12.5} [9]. $\alpha$ and $\beta$ are the weights for the regularization terms. The first term in (13) uses image derivatives for reducing ringing artifacts while the second and third terms are the regularization terms which prefer $L$ with smooth gradients. Note that in our algorithm, we further introduce a variable $v = (v_x, v_y)$ to measure the similarity of $\nabla L$. In this way, the sensitivity of $L$ to the noise can be further reduced. The solution for (13) is:

$$F(L) = \frac{\overline{F(K)} \cdot F(B) \cdot \Delta + \alpha(\overline{F(\partial x)}F(v_x) + \overline{F(\partial y)}F(v_y))}{\overline{F(K)} \cdot F(K) \cdot \Delta + \alpha(\overline{F(\partial x)}F(\partial x) + \overline{F(\partial y)}F(\partial y))} \tag{14}$$

where $F(*)$ denotes the Fourier Transformation, $\overline{F(*)}$ is the conjugate of the Fourier Transformation, and $\Delta = \sum_{\partial_*} \omega_*(\overline{F(\partial_*)} \cdot F(\partial_*))$.

According to the shrinkage formula, we can derive the optimal solution for $v$:

$$(v_x, v_y) = (\frac{\partial_x L}{\|\nabla L\|}, \frac{\partial_y L}{\|\nabla L\|}) \cdot \max(\|\nabla L\| - \frac{\beta}{2\alpha}, 0) \tag{15}$$

From the observation, as the term $\alpha$ becomes smaller, the final image is sharper whereas more noises are produced. Therefore, we propose to handle this problem between sharpening and noise suppression by adaptively adjusting the parameter through the entire iterative process. When the blurred image is downsized at the beginning of iterations, our main concern is to suppress the noises. Otherwise, the effects of noises may degrade in the consecutive iterations especially when the predicted image is up-sampled. And as more iterations are involved, our focus is shifted to enhance the sharpening effect. Therefore, through the iterations at different scales, we adaptively adjust the value of $\alpha$ in the optimization constraint by setting it relatively large in the initial step and letting $\alpha_n = \alpha_{n-1} \cdot \mu$ during iterations at each scale, where $\mu$ is a regularization term which is less than one. With our adaptive scheme, both noise suppression and image sharpening are properly taken into considerations. In our experiments, $\alpha_0$, $\mu$, $\beta$ are set to be 0.2, 0.9, 1, respectively.

Moreover, at each scale, $\lambda$ in (8) is also adjusted decreasingly in each iteration since the predicted image is becoming sharper as the iteration runs at the same scale. So we set $\lambda_0$ equals 1 and let $\lambda_n = \lambda_{n-1} \cdot 0.9$. This scheme works well in our experiment and it is robust among different input blurry images.

## III. EXPERIMENTAL RESULTS

In this section, we show experimental results of our proposed PDE-based deblurring method and our Saliency-based Deblurring (SD) algorithm. We perform two groups of experiments for the uniform motion blur case and spatially-variant blur case, respectively. Note that the parameters $\alpha_0$, $\mu$, $\beta$ in (15) are set to be 0.2, 0.9, 1, respectively throughout the experiments.

### A. *Experiments for Uniform Motion Deblurring*

Our modified deblurring algorithm is effective in the following aspects: (1) PDE based latent image prediction, (2) adaptive deconvolution, and (3) the deblurring results. Based on the above parts, various experiments are conducted.

*1) Latent Image Prediction*

Fig. 4 shows the result of our experiment. In Fig. 4, the blurry image is converted into grayscale. (b) is the result after bilateral and shock filter [9]. And (d) is the result after bilateral filter and our anisotropic PDE edge enhancement method.

Comparing (b) and (d), the effect of our edge prediction approach is apparent. In (b), artifacts are serious around the edges and some edges even appear unnatural. In (c), the white contour is estimated from $-\lambda \Delta I$ in (8) and it exactly represents the true edge, which makes the image in (d) much sharper than the original one without loss of nature.

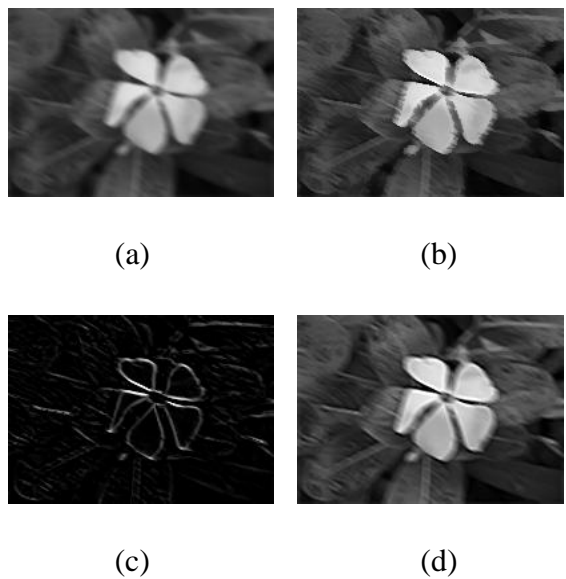

(a)　　　　　　　　(b)

(c)　　　　　　　　(d)

Fig. 4. (a) Original blurry grayscale image, (b) predicted image by shock filter, (c) contour map estimated from PDE, (d) predicted image by anisotropic PDE.

*2) Deconvolution*

To demonstrate the effectiveness of our adaptive deconvolution, we use the same edge prediction scheme with fast deconvolution method in [9] and our adaptive deconvolution, respectively. Fig. 5 shows the deblurring result from [9] (b), the deblurring result by using PDE for edge prediction and then using fast deconvolution in [9] (c), and our deblurring result by using adaptive deconvolution (d). The corresponding estimated kernels are also shown at the right-bottom corners in (b)-(d), respectively.

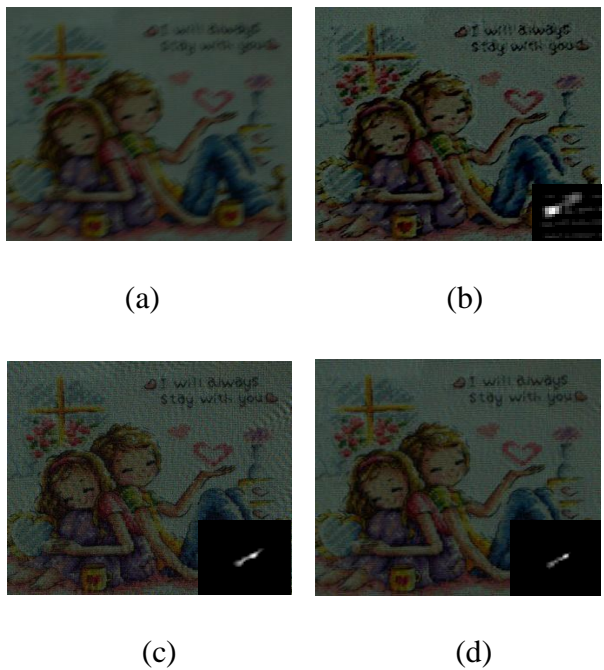

Fig. 5. (a) Original blurry image of 504*500, (b) the deblurring result by [9], (c) the deblurring result by using PDE for edge prediction and then using fast deconvolution in [9], (d) the deblurring result by our PDE-based method (best view in color).

In Fig. 5, the deblurring result directly by [9] (i.e., (b)) is too saturated such that unnatural artifacts are introduced and the words in the top are not clear enough. In (c) where the fast deconvolution mentioned in [9] is adopted, although the image's natural property is preserved, ringing artifacts are serious, which can also be indicated from the kernel. Compared to the above two results, our result is robust. The deblurring effect is significant and artifacts are properly removed as well.

*3) More experimental results*

More experiments are conducted compared with [8], [9], [11] and [12]. The corresponding results are shown in Figs. 6-8.

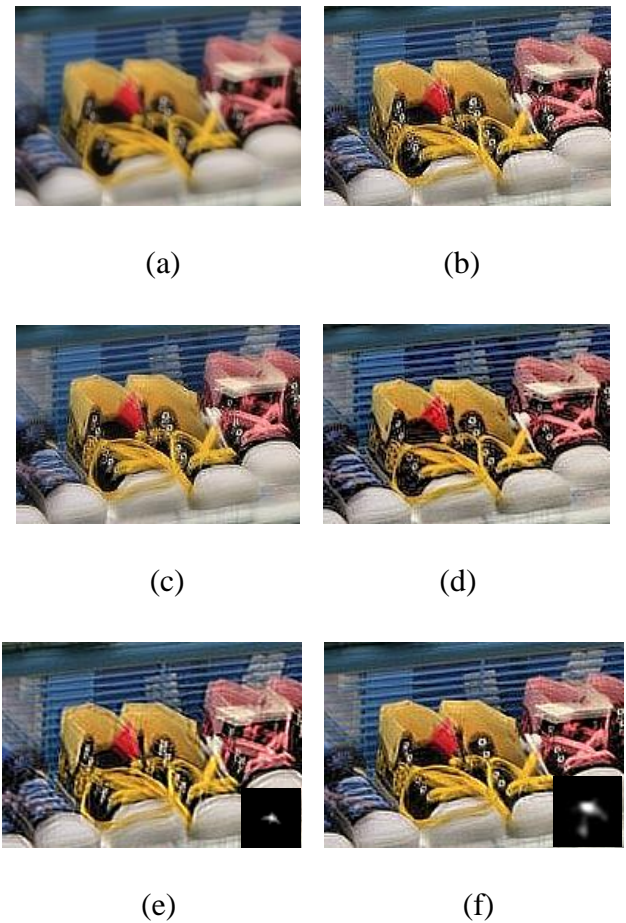

Fig. 6. Deblurring results. (a) part of the original blurry image of 340*280, (b) deblurring result by [9], (c) deblurring result by [8], (d) deblurring result by [11], (e) deblurring result and estimated kernel by [12], (f) deblurring result and estimated kernel by our approach(best view in color).

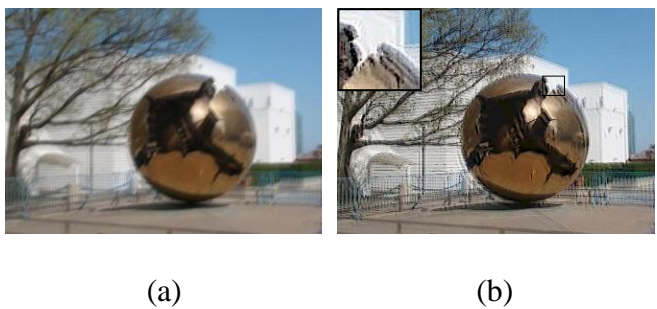

(a)  (b)

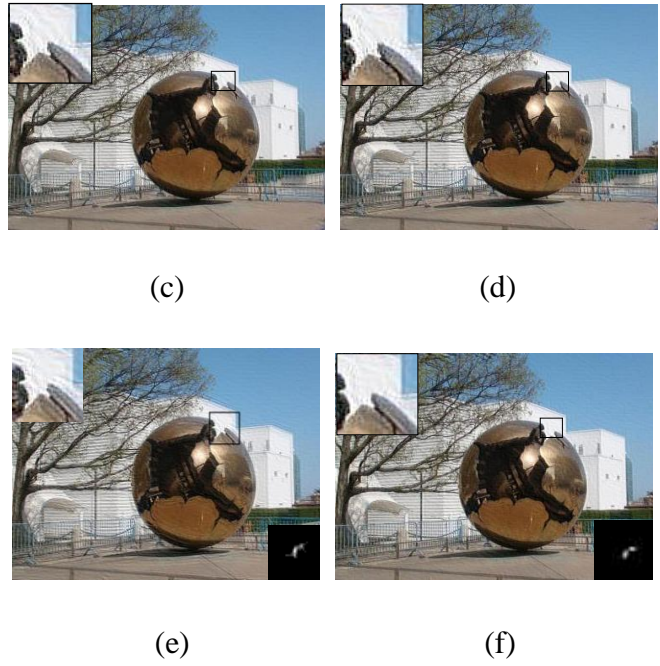

(c)                (d)

(e)                (f)

Fig. 7. Deblurring results. (a) part of the original blurry image of 600*432, (b) deblurring result by [9], (c) deblurring result by [8], (d) deblurring result by [11], (e) deblurring result and estimated kernel by [12], (f) deblurring result and estimated kernel by our approach (best view in color).

From Fig. 6 and Fig. 7, we can see that in the deblurring results by [9], [8] and [12] (i.e. (b), (c) and (e)), the images are not sharp enough and the noises are serious. In the results by [11] (i.e. (d)), the final effects are much sharper but the artifacts are also produced around the edge. With our approach (i.e. (f)), the denoising effects are obviously improved.

In Fig. 8, the blurry images are chosen from real photographs. The results by our proposed PDE-based deblurring method show that the sharp edges have been significantly enhanced and the estimated kernels have reasonable shapes. Besides, almost no ringing artifacts are produced around the arc-shaped boundary.

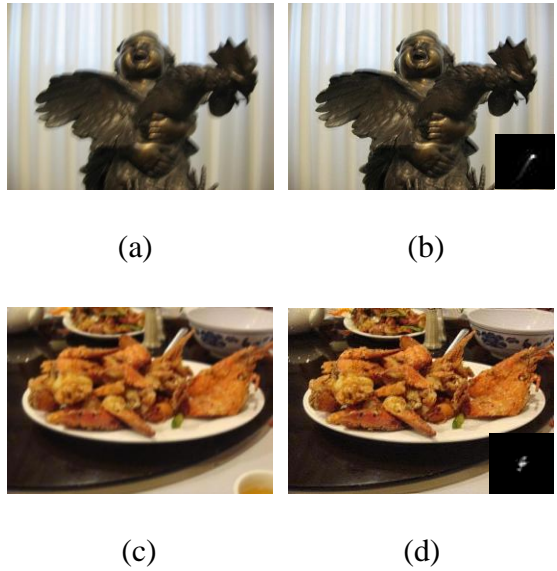

(a) (b)

(c) (d)

Fig. 8. Deblurring results. (a, c) original blurry image (b, d) deblurring results by our method (best view in color).

*4) Quantitative Comparison Results*

In order to test the denoising effectiveness of our approach, we also ran our modified deblurring algorithm over four blurred images including the images from Levin's dataset [14]. In the experiment, we measure Root Mean Squared Errors (RMSE) between the ground truth sharp images and the deblurring results by [8], [9], [11], [12], and our approach.

In the experiment, for each image, a fixed kernel size is used for different deblurring methods. From Table I, the first four columns represent deblurring results performed by [8], [9], [11], and [12]. The rightmost column shows our RMSEs results. We can see that images deblurred by our approach have smaller errors. This further demonstrates that the noises can be more effectively suppressed by our deblurring approach.

Furthermore, Table II compares the time complexity for four blurred images with different sizes on a PC with 2.0 GHz dual-core CPU and 1 G memory. In TABLE II, the first four columns represent the processing time by the executable of [8], [9], [11], and [12]. The last column shows the processing time by our approach with C++ implementation. We can see that the processing time of our method is faster than [8] [11], and [12]. Although the complexity is a little larger than [9], the deblurring results by our method is obviously better than [9]. Note that the complexity of our algorithm can be further reduced after optimization. For example, when our algorithm is implemented with GPU acceleration and other optimization, the average processing time is reduced to less than 5 seconds.

Table I

Comparison of RMSEs between The Ground Truth Sharp Images and The Deblurring Results by [8], [9], [11], [12] and Our Approach

|  | **Deblurring Results by [8]** | **Deblurring Results by [9]** | **Deblurring Results by [11]** | **Deblurring Results by [12]** | **Our Deblurring Results** |
|---|---|---|---|---|---|
| Image 1 | 0.2346 | 0.2391 | 0.2401 | 0.2482 | 0.2310 |
| Image 2 | 0.0590 | 0.0560 | 0.0666 | 0.0619 | 0.0427 |
| Image 3 | 0.0623 | 0.0401 | 0.0411 | 0.0361 | 0.0257 |
| Image 4 | 0.0559 | 0.1121 | 0.0654 | 0.0647 | 0.0474 |

Table II

Comparison of Time Complexity for Algorithms of [8], [9], [11], [12] and Our Approach

| **Image Num** | **Image Size** | **Kernel Size** | **Entire Processing Time (Sec.)** | | | | |
|---|---|---|---|---|---|---|---|
| | | | **[8]** | **[9]** | **[11]** | **[12]** | **Our Method** |
| 1 | 255*255 | 17*17 | 8.47 | 2.34 | 3.408 | 9.33 | 2.86 |
| 2 | 686*508 | 25*23 | 62.59 | 10.98 | 22.22 | 76.47 | 18.30 |
| 3 | 910*754 | 33*32 | 175.12 | 23.70 | 37.00 | 178.26 | 36.62 |
| 4 | 1024*768 | 41*41 | 163.75 | 25.20 | 52.99 | 182.35 | 38.41 |

## B. Experiments for Spatially-variant Blur Case

In this part, the experiments are conducted in four groups: (1) the effectiveness of our Saliency-based Deblurring algorithm, (2) comparison between our approach and other spatial-varying deblurring methods, (3) quantitative comparison results, and (4) more general cases.

### 1) The Effectiveness of our SD algorithm

When the blur is not uniform (i.e., spatially-variant blur problem), the conventional deblurring approaches are not suitable in this case. Based on saliency detection, our SD algorithm can properly handle the problem.

In Fig. 9, the original image has spatially variant blur in which the foreground object is sharp whereas the background clock is blurred. If we estimate the blur kernel from the entire image and use it in the deconvolution step, the deblurring effect is not satisfactory (i.e., (c)). On the other hand, if the background blur kernel is directly applied to the entire image without the compensate method, we can see that artifacts are serious (i.e., (e)). In our SD approach, the saliency map is first calculated. From it, we can see most of the foreground part is extracted as saliency (i.e., (b)). Then we estimate a blur kernel according to the background and perform the compensate method before we use the kernel globally. From (d) and (f), it is apparent that in our method, the background edges are enhanced and the artifacts in the foreground are eliminated in the meanwhile.

In Fig. 10, the original image is from the website which also represents a spatially-variant blur case. In the original image, the foreground athlete is sharp and the background is blurry. If we estimate the kernel globally, it fails to reflect the blur direction due to the uneven blur between foreground and background, thus making the deblurring effects unobvious. (i.e., (c)). If we apply the background kernel to the entire image and fuse the result with the originally sharp saliency part, undesired ringing artifacts will come out around the boundaries of the foreground sharp component for the same reason (i.e., (d)). Compared with these methods, our SD approach first compensates the sharp region such that the entire image is blurred by the same kernel (i.e., (e)). Then, the global deconvolution is performed in our SD approach where the background is deblurred and the foreground sharp part is also preserved (i.e., (f)).

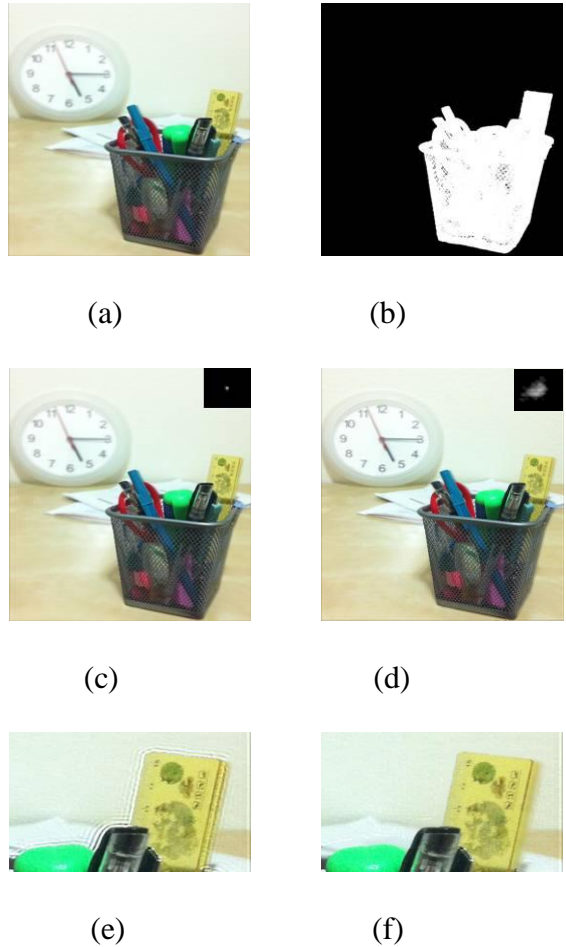

(a)          (b)

(c)          (d)

(e)          (f)

Fig. 9. Spatially-variant deblurring. (a) original image with foreground sharp and background blurred, (b) saliency map of (a), (c) result by global deblurring method, (d) result by our saliency-based deblurring method, (e) zoomed details in (c), (f) zoomed details in (d) (best view in color).

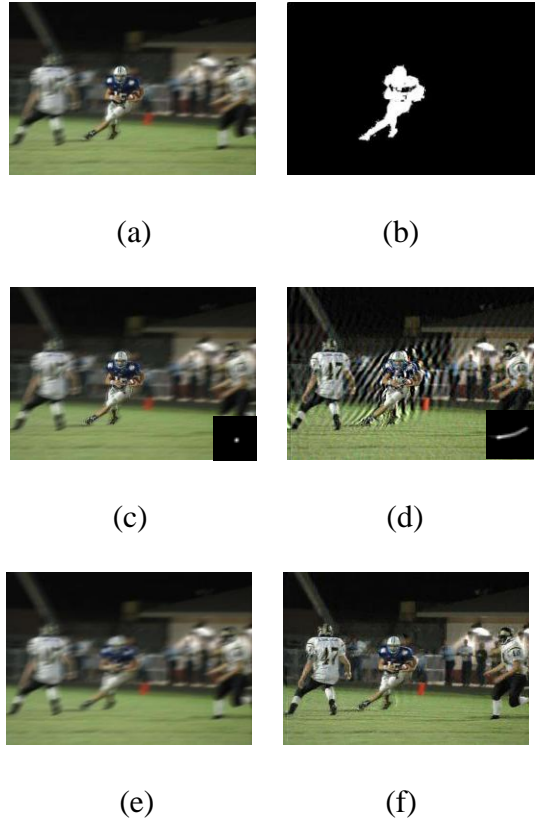

Fig. 10. Spatially-variant deblurring. (a) original image, (b) saliency map of (a), (c) result by global deblurring method, (d) result by applying the background blur kernel to the entire image, (e) compensate fusing result, (f) final output by our SD algorithm (best view in color).

More experiments are conducted to illustrate the results of our SD algorithm. From Fig. 11, we can see that the global deblurring method cannot effectively estimate the blur kernel (i.e., (b)). And applying the background kernel to the entire image without compensation will cause undesired artifacts around boundaries of the sharp foreground components (i.e., (c)). By our SD algorithm, these problems can be solved and the output results are satisfactory (i.e., (d)).

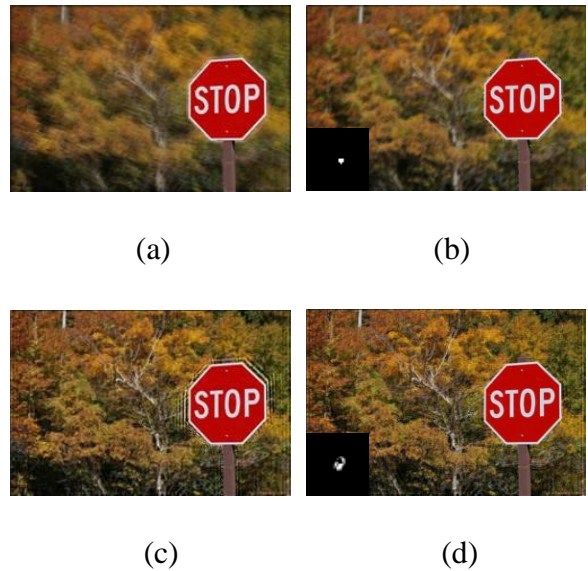

Fig. 11. Spatially-variant deblurring. (a) original image, (b) result by global deblurring method, (c) result by applying the background blur kernel to the entire image, (d) result by our approach (best view in color).

*2) Comparison between Our Approach and Other Spatial-varying Deblurring Methods*

Moreover, we also conduct some experiments comparing the deblurring results of Chan [19] and our approach.

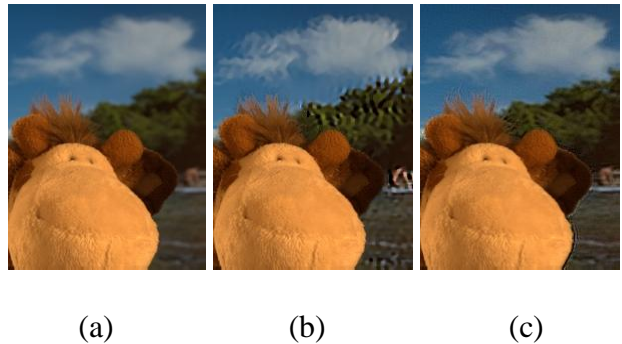

(a) (b) (c)

Fig. 12 Spatially-variant deblurring. (a) original image with foreground focused, (b) deblurring result by Chan [19], (c) our final deblurring result (best view in color).

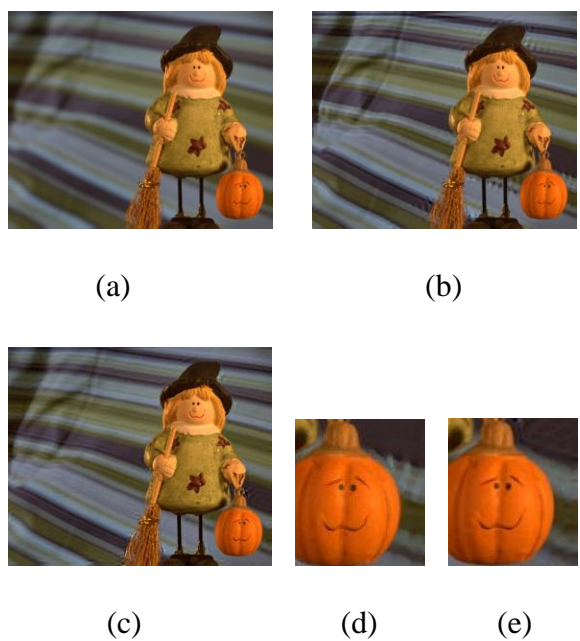

Fig. 13 Spatially-variant deblurring. (a) original image with foreground focused, (b) deblurring result by Chan [19], (c) our final deblurring result, (d) zoomed details in (b), (e) zoomed details in (c) (best view in color).

In Fig. 12 and Fig. 13, the original image is blurred in the background (i.e., (a)). From the deblurring result of [19] (i.e., (b)), we can see that serious artifacts are produced in the background. In the results by our SD algorithm (i.e., (c)), blur in the background is removed whereas artifacts are suppressed as well.

*3) Quantitative Comparison Results*

Table III compares the RMSE and time complexity between the method of [19] and our proposed method over four images from the image matting dataset [24]. The experimental settings are the same as Table I and Table II. From Table III, we can see that our proposed method can achieve better deblurring results (smaller RMSE) than [19] while having similar time complexity with [19]. This further demonstrates the effectiveness of our proposed method. Again, note that the time complexity of our algorithm can be further reduced with GPU acceleration and other optimization, as mentioned in Table II.

Table III

Comparison of RMSEs and Time Complexity between The Deblurring Results by [19] and Our Approach

| Image Number | Image Size | RMSE [19] | RMSE Our method | Time (sec.) [19] | Time (sec.) Our method |
|---|---|---|---|---|---|
| 1 | 800*565 | 0.0217 | 0.0177 | 47.05 | 48.94 |
| 2 | 800*646 | 0.0270 | 0.0241 | 45.03 | 49.28 |
| 3 | 800*580 | 0.0386 | 0.0271 | 49.60 | 45.28 |
| 4 | 800*620 | 0.0207 | 0.0199 | 48.98 | 44.25 |

*4) More General Cases*

In the previous deblurring experiments, for the original blurry image, the foreground component is sharp and the background parts are uniformly blurred. However, as mentioned before, our method can also be applied in more general cases. For example, the foreground is defocused whereas the background is sharp.

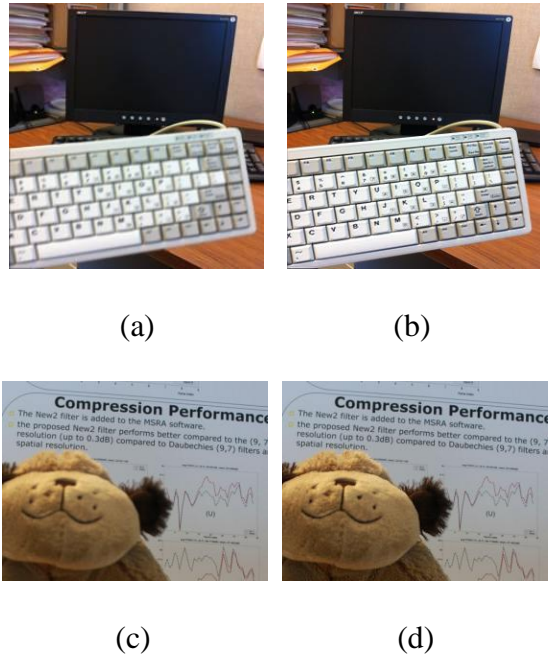

(a)      (b)

(c)      (d)

Fig. 14 Spatially-variant deblurring. (a, c) original spatially variant blur image, (b, d) deblurring result by our approach.

In Fig. 14, the foreground objects are blurred in the original images. In the experiment, after separating the foreground and background information by saliency detection, we perform deblurring algorithm and the compensate method in a similar way. From our results, the deblurring effect is obvious. Besides, almost no artifacts or noises are produced around the boundaries of the blurry components.

Moreover, note that our algorithm can also be easily extended to deblur multiple regions with different blurs (i.e., different regions with different motion blurs in the same image). In this case, we first extract the blurred regions and then apply our deblurring method respectively. For example, assume that there are two blurred foreground objects *A* and *B*. The regions for A, B, and the background components can be formulated as:

$$
\begin{aligned}
Region_A &= MAP_A \cdot Original\_image \\
Region_B &= MAP_B \cdot Original\_image \\
Background &= (1 - MAP_A - MAP_B) \cdot Original\_image
\end{aligned}
\quad (16)
$$

where $MAP_A$ and $MAP_B$ are the binary region masks for the blurred objects *A* and *B*, respectively. Then, we can create two fused images for objects A and B, respectively, as in (17):

$$
\begin{aligned}
Fused\_image_A &= (1 - MAP_A) \cdot (K_A \otimes Original\_image) + Region_A \\
Fused\_image_B &= (1 - MAP_B) \cdot (K_B \otimes Original\_image) + Region_B
\end{aligned}
\quad (17)
$$

where $K_A$ and $K_B$ are the estimated kernels for the blurred regions *A* and *B*. Finally, after we get the deblurred images from the fused images in (17), we can fuse these deblurred images with the sharp background image to create the final debluring result:

$$Final = MAP_A \cdot Deblurred\_image_A + MAP_B \cdot Deblurred\_image_B$$
$$+ (1 - MAP_A - MAP_B) \cdot Original\_image \tag{18}$$

From (16)-(18), we can see that when there are multiple blur regions in an image, we can simply extend our algorithm by processing each region separately. When processing each region, we can view this region as the foreground and the remaining part of the image as the background. Finally, these deblurred regions can be fused to created the final deblurring result. Some experimental results of using our algorithm on multiple blurring regions are shown in Fig. 15. The estimated blur kernels of the blurring regions are also shown in Fig. 15 (b) and (d). Fig. 15 shows that our algorithm is also effective in handling multiple blurring regions.

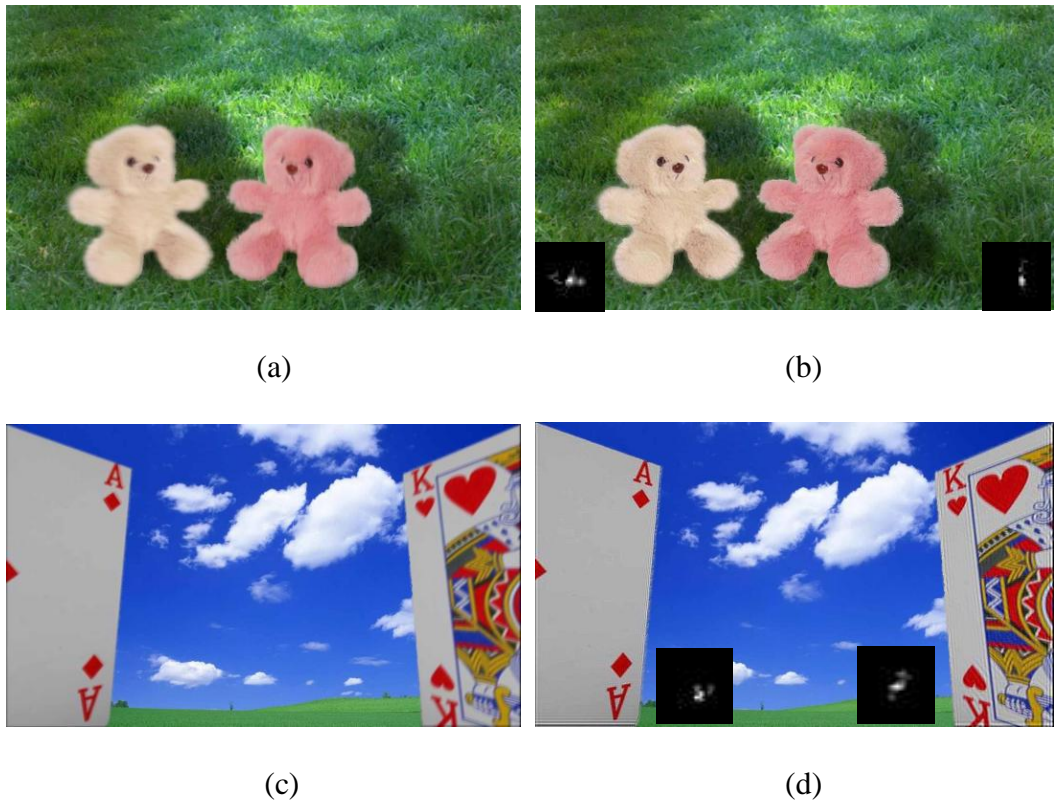

Fig. 15 Debluring images with two blurred foregrounds. (a, c) original blur image, (b, d) deblurring result by our approach (best view in color).

*C. Limitations of the algorithm*

From our experiments, we observe that our algorithm has the following limitations:

Firstly, the saliency detection results may affect the performance of our spatially-variant deblurring algorithm. For example, in Fig. 16, since the saliency detection results in (c) and (f) are less accurate, the deblurring results in (b) and (e) include some unnatural effects (e.g., there are unnatural effects on the coat of the bear in (e)). This problem can be improved by introducing other more accurate region extraction/ segmentation methods or combining them with saliency detection [18, 19, 26]. For example, we can first use saliency detection to achieve the initial salient region, then graph cut segmentation [26] can be applied based on the saliency detection results to achieve a more accurate region extraction result.

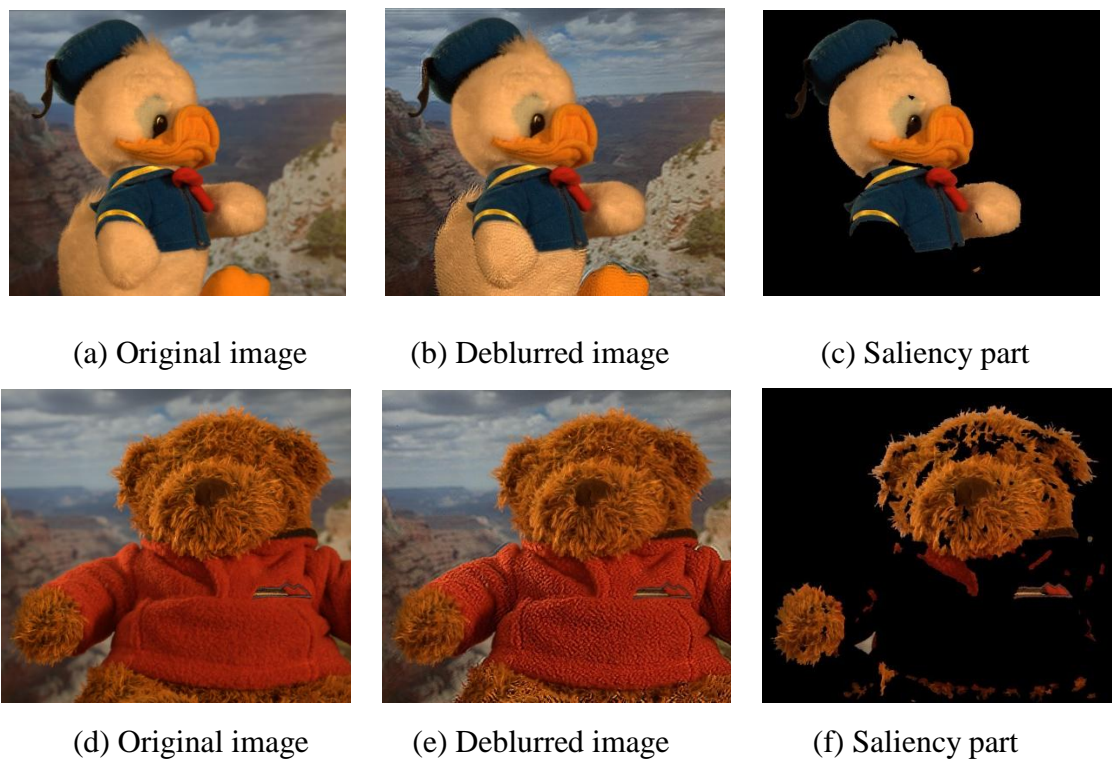

(a) Original image  (b) Deblurred image  (c) Saliency part

(d) Original image  (e) Deblurred image  (f) Saliency part

Fig 16. Results with less accurate saliency detection results.

Secondly, the deblurring results will also be less obvious for small blurry regions (e.g., when the blurred foreground object is very small). This is because when the blurry regions are small, the information is insufficient to achieve reliable blur kernels, making the final results less satisfactory. One possible solution may be developing some method to diffuse the blurry effect to the sharp regions to create a larger "blurry" region for kernel estimation. And this will be one of our future works.

Thirdly, although our algorithm is less sensitive to the parameter settings with the introduction of the adaptive deconvolution scheme, the results by our algorithm are still relatively varying with different parameter values. This is a common problem for most of the existing deblurring algorithms [8-16]. And it will be another part of our future works to further improve the robustness to parameters or develop new schemes to adaptively decide the optimal parameters.

## IV. CONCLUSION

In this paper, we propose a new algorithm to effectively address the spatially variant blur problem. Firstly, the proposed algorithm addresses the spatially variant blur problem by introducing a Saliency-based Deblurring (SD) method and a compensate approach such that the blur is locally removed whereas sharp components are preserved. Secondly, we also propose a PDE-based deblurring method which adopts anisotropic PDE model for edge prediction as the initial step to estimate PSF. Thirdly, we also employs an adaptive optimization constraint for kernel estimation and deconvolution based on image derivatives. Experimental results demonstrate effectiveness of the proposed algorithm by comparing with the state-of-the-art algorithms.